# A Novel Multi-Objective Evolutionary Algorithm for Counterfactual Generation

Gabriel Doyle-Finch and Alex A. Freitas

gabrielsaul.commerce@outlook.com, A.A.Freitas@kent.ac.uk

School of Computing, University of Kent
Canterbury, CT2 7FS, United Kingdom

**Abstract**
Machine learning algorithms that learn black-box predictive models (which cannot be directly interpreted) are increasingly used to make predictions affecting people's lives. It is important that users understand such models' predictions, particularly when the model outputs a negative prediction for the user (e.g. denying a loan). Counterfactual explanations provide users with guidance on how to change some of their characteristics to receive a different, positive classification by a predictive model. For example, if a predictive model rejected a user's loan application, a counterfactual explanation might state: "If your salary was £50,000 (rather than your current £35,000), then your loan would be approved." This paper proposes two novel contributions: (a) a novel multi-objective Evolutionary Algorithm (EA) for counterfactual generation based on lexicographic optimisation, rather than the more popular Pareto dominance approach; and (b) an extension to the definition of the objective of validity for a counterfactual, based on measuring the resilience of a counterfactual to violations of monotonicity constraints which are intuitively expected by users – e.g., intuitively, the probability of a loan application to be approved would monotonically increase with an increase in the applicant's salary. Experiments involving 15 experimental settings (3 types of black box models times 5 datasets) have shown that the proposed lexicographic optimisation-based EA is very competitive with an existing Pareto dominance-based EA; and the proposed extension of the validity objective has led to a substantial increase in the validity of the counterfactuals generated by the proposed EA.

## 1 Introduction

This work focuses on the classification task of supervised machine learning [Mitchell 1997], where the input data consists of instances (examples) with two types of variables: a set of predictive features and a class variable. The goal of a classification algorithm is to learn from data a model to predict the class label of an instance (e.g., whether a loan application is granted or denied) based on the values of the predictive features for that instance (e.g. characteristics of the person applying for a loan, like their age, salary, etc). The algorithm learns a classification model from a training dataset, where the class labels of the instances are known, and then the model can be used for predicting the class labels of instances in a testing dataset, where the class labels of the instances are unknown.

State-of-the-art classification algorithms often learn classification models with a very high predictive accuracy, but their learned models are usually "black-box" models as it is impossible for a person to understand the model's internal logic, unless a post-processing method is used for explaining the model or its specific predictions (class labels) [Burkart & Huber 2021]. With the large increase in the use of classification algorithms across multiple application domains, the ability to explain the predictions of a classification model is crucial for garnering trust on machine learning and helping users navigate machine learning applications with less uncertainty. For example, the requirement for machine learning explainability remains one of the biggest challenges in medicine [Holzinger et al. 2019] and shows high demand in the financial sector [Heng & Subramanian 2022].

Among the various types of methods for explaining black-box models' outputs that have been developed in the literature [Burkart & Huber 2021]; this work focuses on one of the most promising methods, *counterfactual explanation* [Guidotti 2024],[Stepin et al. 2021], which tells a user how to alter an instance's feature values to attain a different classification from the original classification given by the black-box model. For example, if a black-box model predicts that a loan application should be denied, a counterfactual explanation could recommend that the applicant increase their salary to a certain amount, which would be estimated as a sufficient condition for the black-box model to predict that the loan would be granted. Hence, unlike other types of black-box model explanations in general, counterfactual explanations have the important advantage of being designed to be directly actionable, by recommending how feature values in an instance should be changed in order for the black-box model to predict a desirable class label (e.g., "granted loan") for that instance, rather than predicting an undesirable class label (e.g. "denied loan") for the original instance's feature values.

Since the first method of producing counterfactual explanations was proposed [Wachter et al. 2017], many such methods were proposed, often focusing on optimising a particular objective (counterfactual quality measure), e.g.: focusing on producing more *actionable* counterfactual explanations [Ustun etal. 2019],[Guidotti et al. 2024]; generating *diverse* counterfactual explanations [Russel 2019]; reducing the *magnitude of the feature-value changes* suggested by a counterfactual [Sharma et al. 2020]; or reducing the *number of features* whose values are changed by a counterfactual [Schleich et al. 2021]. In reality, the quality of a counterfactual depends on multiple criteria [Guidotti 2024], and hence a promising research direction consists of optimising multiple counterfactual quality criteria at the same time, using multi-objective optimisation methods [Emmerich & Deutz 2018],[Freitas 2004].

Hence, Evolutionary Algorithms (EAs) are a promising method for counterfactual generation, as EAs are particularly suitable for multi-objective optimisation [Deb 2002],[Emmerich & 2018], due to two main reasons. First, EAs evolve a population of candidate solutions, with different solutions representing different trade-offs among the multiple objectives. Second, EAs compute the fitness (quality) of each complete solution during their search (instead of evaluating partially constructed solutions) and they don't require knowledge of the derivative of any of the objectives being optimised; which facilitates the use of multi-objective fitness functions, including a mixture of continuous and discrete objective functions. Indeed, several EAs for counterfactual generation have been recently proposed (see Section 2.3); of which the most relevant for this work is the MOC (Multi-Objective Counterfactuals) EA proposed in [Dandl et al. 2020], which uses the concept of Pareto dominance to optimise multiple properties of a counterfactual.

In this context, this paper proposes two novel contributions, as follows. First, we propose a novel type of multi-objective EA for counterfactual generation based on lexicographic optimisation, rather than based on Pareto dominance (see Section 2.2 for a review of lexicographic optimisation and Pareto dominance as multi-objective optimisation methods). Second, we propose an extension to the definition of the objective of validity, which is the most important objective (counterfactual property) being optimised by the EAs in this work. This extension is based on measuring the resilience of a counterfactual to violations of monotonicity constraints [Cano et al. 2019], which are constraints intuitively expected by users – e.g. users would intuitively expect that the probability of a loan application to be approved would monotonically increase with an increase in the applicant's salary. Unfortunately such intuitive constraints are not automatically enforced by existing counterfactual-generation methods in general, hence the relevance of this contribution.

The proposed EA is compared with the MOC EA [Dandl et al. 2020] in 15 experimental settings – consisting of learning three types of black-box models (neural networks, random forests and Support Vector Machines (SVM)) from each of five different classification datasets. Overall, the computational results have shown that the proposed lexicographic optimisation-based EA is very competitive with the Pareto dominance-based MOC EA, with the former obtaining in particular clearly better results regarding the optimisation of the validity objective of the generated counterfactuals, which is the most fundamental property of a counterfactual; whilst the proposed EA and MOC are, broadly speaking, equally effective in the optimisation of the three other objectives that make up their multi-objective fitness function. In addition, the proposed extension of the validity objective has led to a substantial increase in the validity of the counterfactuals generated by the proposed EA, and also to a small increase in the validity of the counterfactuals generated by MOC.

The remainder of this paper is structured as follows. Section 2 reviews the background on counterfactual explanations and multi-objective optimisation, as well as related work on EAs for counterfactual generation. Section 3 describes the proposed multi-objective EA for counterfactual generation and the proposed validity objective's extension based on the concept of monotonicity constraints. Section 4 describes the experimental setup for evaluating the proposed methods. Section 5 presents and discusses the computational results. Section 6 concludes with a summary of this work's contributions and avenues for future work.

## 2 Background and Related Work

### 2.1 Counterfactual Explanations

Counterfactual explanations are provided to users as a path of recourse alongside machine learning classifications, i.e. characteristics the user should change to achieve a different predicted outcome. Let $f$ denote a classification model and $\hat{f}(x_{pt})$ denote the class predicted by $f$ for a data point (instance) of interest $x_{pt}$. A counterfactual explanation suggests how some feature(s) of $x_{pt}$ need to be changed for $f$ predicting a different class label for $x_{pt}$. For example, if $f$ rejects a person's loan application, i.e. $\hat{f}(x_{pt})$ = "rejected", a counterfactual explanation would specify how to change some feature(s) of $x_{pt}$ to achieve $\hat{f}(x_{pt})$ = "approved". A counterfactual explanation for this example could be: "If your salary was £50,000 (rather than your current £35,000), your loan would be approved".

We can distinguish between the terms "counterfactual" and "counterfactual explanation'". A counterfactual is a data point $x_{cf}$ constructed by applying minimal changes $\delta$ to the features of a data point of interest $x_{pt}$ that, when inputted to a predictive model $\hat{f}(x_{pt})$, leads to a change in the model's output for $x_{pt}$ [Guidotti 2024]. Thus, a data point $x_{cf}$ is counterfactual if it satisfies the following condition: $x_{cf} = \delta x_{pt}$ AND $\hat{f}(x_{cf}) \neq \hat{f}(x_{pt})$. A counterfactual explanation is the plain language statement that clearly and succinctly provides guidance on the changes $\delta$ from the original data point of interest $x_{pt}$ to the counterfactual $x_{cf}$. For the remainder of this paper, we will refer mainly to counterfactuals, since they have a more succinct representation and once a counterfactual has been generated, the corresponding counterfactual explanation (in plain language) can be easily generated.

Counterfactuals have properties and adhere to certain constraints. The *sparsity* of a counterfactual $x_{cf}$ is number of features with changed values in $x_{cf}$, by comparison with the corresponding feature values in $x_{pt}$. The *distance* between $x_{cf}$ and $x_{pt}$ measures the magnitude by which these features are changed in $x_{cf}$, by comparison with the feature values in $x_{pt}$. In addition, a counterfactual exhibits adherence to the constraint of *actionability*, suggesting changes in feature values that a user can realistically manage in real life – e.g., a user can reduce their debt, but not their age. Counterfactual properties will be discussed in more detail in Section 2.3.

### 2.2 Multi-Objective Optimisation

The most common and simplest approach to cope with multiple objectives being optimised is to assign a numeric weight to each objective, reflecting its importance, and then optimise the weighted sum of the objectives. In effect, this converts a multi-objective optimisation problem into a simpler single-objective optimisation problem, but it has significant disadvantages: the objectives' weights are usually largely arbitrary values, specified in an ad-hoc fashion; and the optimiser considers just one possible trade-off among the objectives [Freitas 2004],[Gonzales et al. 2021].

Hence, this weighted-sum approach is no longer considered in this work; and we briefly review next two more principled multi-objective optimisation approaches that largely avoid the disadvantages of the weight-sum approach, namely: the lexicographic optimisation and the Pareto dominance approach. Both these approaches consider many possible trade-offs among multiple objectives being optimised, in a single run of the optimiser.

#### 2.2.1 The lexicographic multi-objective optimisation approach

The lexicographic optimisation approach consists of optimising objectives in a user-defined decreasing order of priority. More precisely, when two or more candidate solutions are compared to select the best one, they are first compared regarding the first (highest-priority) objective. If one solution is "substantially better" (defined more precisely below) than the other(s) regarding that objective, that former solution is selected. Otherwise, the solutions are compared regarding the second objective. Again, if one solution is substantially better than the other(s), that former solution is selected; otherwise the solutions are compared regarding the third objective, etc. This process proceeds until an outright winning solution is found or until all objectives are considered. In the latter case, if there was no clear winner, we can select the solution with the best value of the highest-priority objective (even if it is just slightly better, rather than substantially better than the other solutions).

In general, a solution is considered "substantially better" than another when their quality difference is above a small user-defined "tolerance threshold", e.g. 0.01 when objectives are normalised into the

[0..1] range. The idea is that very small differences in objective values (below the tolerance threshold) count as "ties", rather than being used for selecting the best solution.

The main advantage of the lexicographic approach is that it allows the optimiser to consider user-defined priorities for the different objectives in a simple and principled way [Freitas 2024],[Zhang et al. 2023], instead of using ad-hoc, largely arbitrary weights for objectives as in the weighted sum approach. Note that it is in general much easier and more natural for users to specify a *qualitative* ordering of objectives' priorities than to specify precise numeric weights for objectives. For example, in the classification task, there is a general consensus that the objective of minimising the prediction error of a classification model has higher priority than the objective of minimising the size of the model; but it would be hard for users to find a good justification to assign to those two objectives specific weights like 0.8 vs 0.2, 0.67 vs 0.33, or any other arbitrary pairs of weights.

The main disadvantage of the lexicographic approach is that it introduces a new user-defined parameter, the tolerance threshold – which is not needed in the Pareto approach discussed later. However, in practice this parameter is simply set to a small value, typically in the range of 0.01 to 0.05 (on a normalised scale from 0 to 1 for the objective values), which is still simpler and more natural to users than specifying the objectives' weights in the weighted-sum approach [Freitas 2024]. In addition, a potential limitation of the lexicographic approach is that, for some sets of objectives (depending on the user's interests), the user's choice of a priority ordering for the objectives might not be so natural, or perhaps only part of the ordering is natural. Actually, in this work, where four objectives (counterfactual properties) will be optimised, two of the objectives have a more natural priority order, but the ordering for the other two objectives is less natural, and so we will do experiments with two different priority orderings, as will be discussed in more detail later.

The lexicographic optimisation approach has been previously used in several machine learning tasks, like feature selection [Gonzales et al. 2021],[Jungjit & Freitas 2015], decision tree induction [Basgalupp et al. 2009], and hyperparameter optimisation for classification algorithms [Zhang et al. 2023]. However, the proposed lexicographic optimisation-based EA (detailed in Section 3.1) is, to the best of our knowledge, the first lexicographic optimisation-based EA for counterfactual generation.

### 2.2.2 The Pareto dominance approach for multi-objective optimisation

The basic idea of this approach is that a candidate solution is Pareto-dominant over another solution if the former satisfies the following two criteria [Deb 2002],[Emmerich & Deutz 2018],[Freitas 2004]: (a) the dominant solution performs at least as well as the dominated solution regarding all objectives; and (b) the dominant solution performs better than the dominated solution regarding one or more objectives. Nondominated solutions are solutions that are not dominated by any other solutions. The set of nondominated solutions is called the Pareto front.

The concepts of Pareto dominance and Pareto front are illustrated in Figure 1, where the two objectives to be minimised are a predictive model's error rate and the model's size. In this figure, the green circles and the yellow triangles represent non-dominated and dominated candidate solutions, respectively. The Pareto front consists of all non-dominated solutions. Consider for example the candidate solution $x_1$. This solution dominates the candidate solution $x_3$, because $x_1$ has smaller values than $x_3$ for both objectives. Solution $x_1$ also dominates solution $x_2$, because although they have the same value for the error rate objective, solution $x_1$ has a smaller value for the model size objective.

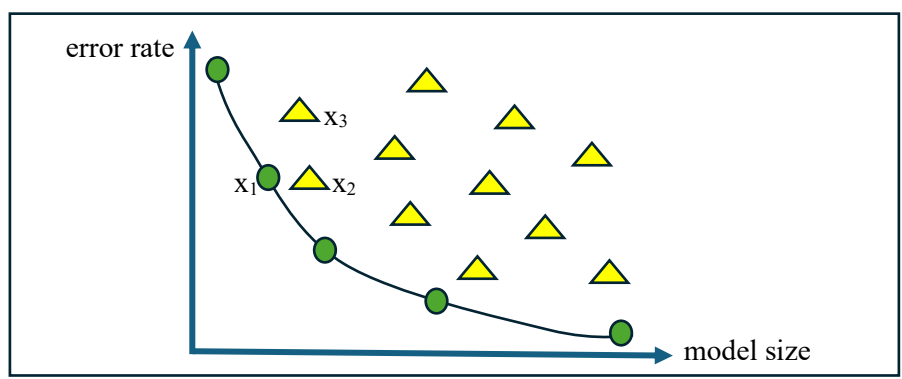

**Figure 1:** Example of Pareto dominance: the candidate solutions represented by green circles make up the Pareto front, whilst the candidate solutions represented by yellow triangles are dominated

The main advantage of the Pareto-dominance approach is that it does not require any user-specified parameter (unlike the weighted-sum and lexicographic optimisation approaches). The main disadvantage of the Pareto approach is that it does not incorporate any notion of higher priority for objectives that are more important to users (unlike the lexicographic approach) [Freitas 2024]. In the above example of minimising a model's prediction error and size, the Pareto approach would be unable to recognise that minimising prediction error is more important. Hence, when using an EA based on Pareto dominance, a candidate solution with large error but minimal model size would tend to be nondominated and remain in the population of an EA for many generations, even though such a solution would not be acceptable to a user. Another issue is that an optimiser based on the Pareto approach usually returns a large set of nondominated solutions. It is usually assumed that a user can select the best nondominated solution based on their preferences, but this selection may not be easy for users. These points will be further discussed in Section 3.1, in the context of EAs.

**2.3 Related Work on Evolutionary Algorithms (EAs) for Counterfactual Generation**
This subsection briefly reviews EAs for generating counterfactuals. For a more comprehensive discussion of many types of methods for counterfactual generation, the reader is referred to [Guidotti 2024],[Stepin et al. 2021]. Counterfactuals can be generated for any type of classification model, but in general counterfactuals are particularly valuable to explain the predictions of a black-box model, i.e. a model that does not provide a user-understandable explanation for its predictions, e.g. a neural network or a support vector machine model. Hence, in the remainder of this paper we assume that a black-box classification model has been learned from a dataset where each instance represents an object to be classified. Each instance comprises a set of features (attributes) describing properties of that object (e.g. the age, salary, etc, of a customer) and a class variable (e.g. credit rating). The dataset is divided into a training set and a testing set. The black-box model is learned from the training set, and it is then used to predict the value of the class attribute for instances in the testing set.

An EA-based counterfactual generator takes as input a black-box classification model and a data point of interest $x_{pt}$, which is an instance in the testing set. I.e., $x_{pt}$ is the instance whose prediction will be explained by a counterfactual. The black-box model will be used to compute part of the fitness function of the EA, more precisely to test the validity of a counterfactual (candidate solution), as discussed later. The fitness function can also measure several other properties of counterfactuals in a multi-objective scenario, as discussed later.

Candidate solutions are mutated forms $\delta x_{pt}$ of the original data point of interest $x_{pt}$, and the EA aims to generate valid counterfactuals, satisfying the inequality: $\hat{f}(\delta x_{pt}) \neq \hat{f}(x_{pt})$ – i.e., the black-box model predicts different class labels for the candidate counterfactual $\delta x_{pt}$ and the point of interest $x_{pt}$. The fitness function of the EA could constitute a single objective, e.g. minimise the distance between $\delta x_{pt}$ and $x_{pt}$, or multiple objectives (the approach taken in this work).

Figure 2 shows an example of the generation of two EA individuals representing candidate counterfactuals by mutating an original point of interest ($x_{pt}$). In this simple example, there are only four features, each corresponding to a gene in an individual of the EA population, and the class variable indicates whether a person's loan application is denied or granted. The features' meanings are: a person's age, gender, salary and whether or not they own a house (own-house), and the features' values for $x_{pt}$ are shown in the left part of the figure. The right part of the figure shows two EA individuals generated from $x_{pt}$, and their probability of the "granted" class predicted by a black-box model, denoted $\hat{f}_p(x_{pt})$. Note that we use the symbol $\hat{f}_p()$ to denote the *probability* of the target class output by a black-box model, which is different from the symbol $\hat{f}()$ used to denote the *class* output by a black-box model. In this figure, the first individual was generated by mutating the age and own-house features; whilst the second individual was generated by mutating the salary feature (increasing the person's salary from £35K to £42K). The probability of the "granted" class computed by a black-box model for these two individuals are 0.2 and 0.7, respectively, leading to the predicted classes "denied" and "granted". Since the target class is "granted", only the second (bottom) of these two EA individuals is considered a valid counterfactual.

The remainder of this section reviews three relatively recent EAs for counterfactual generation, as the most related work to the EA proposed in this paper.

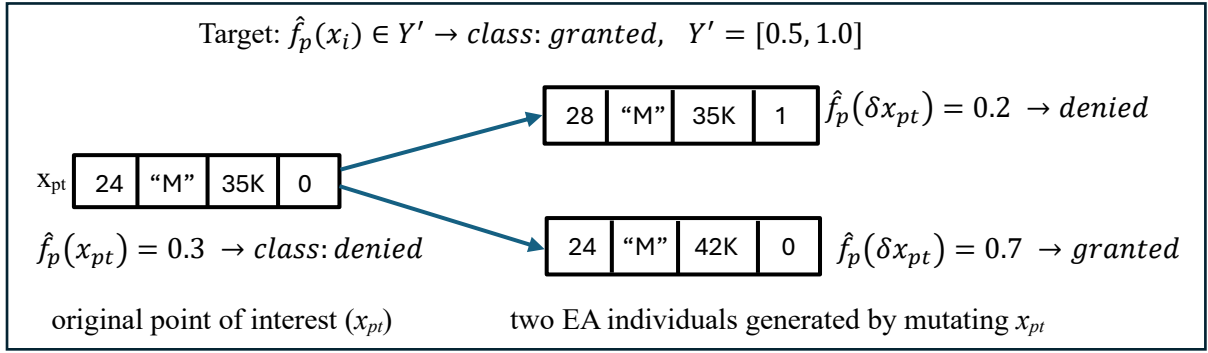

**Figure 2:** Example of the generation of two candidate solutions (individuals in the EA population), which are generated by mutating the original point of interest (a test instance)

### 2.3.1 CERTIFAI

Counterfactual Explanations for Robustness, Transparency, Fairness and Interpretability of Artificial Intelligence models (CERTIFAI) [Sharma et al. 2020] is a single-objective EA-based counterfactual generator designed to address common issues with black-box predictive models, including *robustness*. Robustness is the ability of a predictive model to generalise well and maintain accuracy when presented with new data. A robust model is not oversensitive to perturbations in the input data, which will reduce the misclassification of data points that differ only slightly from correctly classified examples [Goodfellow et al. 2015]. Adversarial attacks on predictive models [Nguyen et al. 2015],[Carlini & Wagner 2017] can discover blind spots in the models by applying minimal perturbations to inputted data points and exploiting the observed changes in output produced by highly similar data points. CERTIFAI demonstrates the use of counterfactuals for assessing a model's vulnerability to such attacks by evolving adversarial examples [Szegedy et al. 2014] of correctly classified data points. Adversarial examples can be considered counterfactuals that are near-identical to the original data points of interest.

CERTIFAI's EA uses a single-objective fitness function to evaluate candidate solutions (counterfactuals) that are similar to the original data point of interest $x_{pt}$ by returning a higher (better) fitness value for individuals that are a minimal distance from $x_{pt}$, based on the following formula:

$$fitness(\delta x_{pt}) = 1/(dist(x_{pt}, \delta x_{pt})),$$

where *dist*() is a given distance function. Considering the performance of a solution regarding a single objective is simple, but it fails to consider other counterfactual properties, such as sparsity.

### 2.3.2 GeCo

Genetic Counterfactuals (GeCo) [Schleich et al. 2021] is a multi-objective EA-based counterfactual generator. It uses the following fitness function, based on validity (the condition $\hat{f}_p(\delta x_{pt}) > 0.5$, where $\hat{f}_p(\delta x_{pt})$ is the black-box model's predicted probability of the target class for a candidate counterfactual $\delta x_{pt}$) and the distance between a point of interest $x_{pt}$ and $\delta x_{pt}$ ($dist(x_{pt}, \delta x_{pt})$):

$$fitness(\delta x_{pt}) = \begin{cases} dist(x_{pt}, \delta x_{pt}), & \text{if } \hat{f}_p(\delta x_{pt}) > 0.5 \\ (dist(x_{pt}, \delta x_{pt}) + 1) + (1 - \hat{f}_p(\delta x_{pt})), & \text{otherwise} \end{cases}$$

i.e., the fitness of $\delta x_{pt}$ is given by $dist(x_{pt}, \delta x_{pt})$ if it is a valid counterfactual (i.e., it predicts the target class); otherwise the fitness is a combination of $dist(x_{pt}, \delta x_{pt})$ and how close $\hat{f}_p(\delta x_{pt})$ is to the minimum desirable value of 0.5, which penalises invalid candidate solutions. The distance function $dist(x_{pt}, \delta x_{pt})$ is a weighted sum of three objectives to be minimised: (a) sparsity, i.e., the number of features changed in $\delta x_{pt}$, by comparison with the feature values in $x_{pt}$; (b) the average distance (feature-value change) between $\delta x_{pt}$ and $x_{pt}$ across all features; (c) the maximum feature value change across all features. These objectives are weighted by three parameters: α, β, γ, where α + β + γ = 1.

Note that the choice of values for the weights *α, β, γ* is in principle arbitrary or based on a vague intuition of the user, which is a typical characteristic of the weighted-sum approach for multi-objective optimisation [Freitas 2004]. Hence, the chosen weight distribution can lead to an undesirable result for the user. For example, in an experiment in [Schleich et al. 2021], increasing α resulted in each counterfactual having a single feature changed by a large distance, whilst other counterfactuals in the search space might be better solutions due to requiring less overall change across many features. This supports the importance of considering a more principled multi-objective optimisation approach, like the lexicographic optimisation or Pareto dominance approaches.

### 2.3.3 Multi-Objective Counterfactuals (MOC)

MOC [Dandl et al. 2020] is a counterfactual-generator EA that performs multi-objective optimisation based on Pareto dominance. MOC's multi-objective fitness function consists of four objectives, viz.:

**Validity (objective $o_1$):** A mutated point of interest $\delta x_{pt}$ is a valid counterfactual if and only if it is classified by the model differently with respect to the original point of interest $x_{pt}$ [Guidotti 2024], i.e. $\hat{f}(\delta x_{pt}) \neq \hat{f}(x_{pt})$. MOC calculates validity as:

$$o_1(\hat{f}_p(\delta x_{pt}), Y') = \begin{cases} 0, & \text{if } \hat{f}_p(\delta x_{pt}) \in Y' \\ \inf_{y' \in Y'} |\hat{f}_p(\delta x_{pt}) - y'|, & \text{else} \end{cases}$$

where $\hat{f}_p(\delta x_{pt})$ is the model's predicted probability of the target class for $\delta x_{pt}$ and *Y'* = [0.5, 1] is the target range for that probability. So, $\delta x_{pt}$ is a counterfactual if $0.5 \leq \hat{f}_p(\delta x_{pt}) \leq 1$. If $\delta x_{pt}$ is not a counterfactual, then the absolute value of the difference between the predicted probability and the smallest *y'* (0.5) in the target range is returned – e.g., if $\hat{f}_p(\delta x_{pt}) = 0.35$ then 0.15 is returned. Hence, the range of values returned by $o_1$ is [0, 0.5], where 0 indicates a valid solution (counterfactual) and any other value in that range indicates the difference between the model's predicted probability of the target class for $\delta x_{pt}$ and the lower bound 0.5 for the target range *Y'*.

**Distance to point of interest (objective $o_2$):** The distance between a mutated point of interest $\delta x_{pt}$ and the original point of interest $x_{pt}$, denoted *dist*($\delta x_{pt}$, $x_{pt}$), should be minimal [Guidotti 2024]. MOC calculates this objective as:

$$o_2(\delta x_{pt}, x_{pt}) = \frac{1}{p} \sum_{i=1}^{p} dist(\delta x_{pt}^{[i]}, x_{pt}^{[i]})$$

where *p* is the number of features, [i] denotes the *i*-th feature of $\delta x_{pt}$ or $x_{pt}$, and *dist*() is the Gower distance function, which handles both numeric and categorical features, as follows:

$$dist\left(\delta x_{pt}^{[i]}, x_{pt}^{[i]}\right) = \begin{cases} \frac{1}{\hat{V}^{[i]}} \left|\delta x_{pt}^{[i]} - x_{pt}^{[i]}\right| & \text{if } x_{pt}^{[i]} \text{ is numerical} \\ I\left(\delta x_{pt}^{[i]} \neq x_{pt}^{[i]}\right) & \text{if } x_{pt}^{[i]} \text{ is categorical} \end{cases}$$

where $\hat{V}^{[i]}$ is the value range (difference between maximum and minimum values) of the *i*-th feature in the observed training set and *I*() is an indicator function that returns 1 if its argument is true and 0 otherwise. The range of values returned by $o_2$ is normalised into [0, 1], where values closer to 0 indicate candidate better solutions (more similar to the original point of interest $x_{pt}$).

**Sparsity (objective $o_3$):** A mutated point of interest $x_{pt}$ is sparse if the number of features whose values changed from the original point of interest $x_{pt}$ is minimal [Guidotti 2024]. MOC calculates sparsity as:

$$o_3(\delta x_{pt}, x_{pt}) = \sum_{i=1}^{p} I\left(\delta x_{pt}^{[i]} \neq x_{pt}^{[i]}\right)$$

where the terms in this equation are the same as used for defining objective $o_2$. The range of integer values returned by $o_3$ is {0, …, *p*}.

**Plausibility (objective $o_4$):** A mutated point of interest $\delta x_{pt}$ should be plausible in the sense that the distance between the feature values of $\delta x_{pt}$ and the feature values of the nearest neighbour(s) of $\delta x_{pt}$ in the training set $X$ should be small [Guidotti 2024]. MOC calculates plausibility as:

$$o_4(\delta x_{pt}, X) = \frac{1}{p}\sum_{i=1}^{p} dist(\delta x_{pt}^{[i]}, x_{pt}^{[i]})$$

This objective measures the Gower distance between $\delta x_{pt}$ and its nearest instance in the training set $X$, to determine if $\delta x_{pt}$ could have plausibly originated from $X$. The range of values returned by $o_4$ is normalised into [0, 1], where values closer to 0 are better, as they indicate greater plausibility.

MOC also enforces two constraints when creating new candidate solutions (counterfactuals) during the EA's run: (a) *Actionability*: MOC can receive a user-defined set of nonactionable features as input, and these features undergo no changes through the EA's run (e.g. setting "gender" as a nonactionable feature prevents the EA from creating a counterfactual suggesting that a person changes their gender); and (b) *Feasibility*: When mutating feature values to create new candidate solutions, any new feature value must be no less than the minimum and no more than the maximum observed value for that feature in the training dataset.

MOC's EA is based on the well-known NSGA-II algorithm [Deb et al. 2002], which uses a nondominated selection function wherein candidate solutions compete for Pareto dominance. In addition, when selecting among nondominated solutions, a crowding procedure is used to favour more diverse solutions (individuals), i.e. solutions that are more distant (different) from other solutions. Upon the termination of the EA's run, a diverse set of solutions is returned, though it is not guaranteed that all of these are valid counterfactuals. Furthermore, while multiple solutions can offer flexibility of choice to the user, returning a large number of solutions to the user can make it difficult for them to ultimately choose a single solution for decision-making in a real-world scenario [Freitas 2024], and so returning a single high-quality solution might be preferable in some scenarios. These issues are addressed in the EA proposed in this paper, based on the lexicographic optimisation approach (rather than Pareto dominance approach) for multi-objective optimisation, as described in Section 3.

## 3 The Proposed Multi-Objective Evolutionary Algorithms for Counterfactual Generation

The main contribution of this paper is to propose a novel multi-objective EA for generating counterfactuals used as explanations for the predictions of black-box classifiers, based on the lexicographic optimisation approach (reviewed in Subsection 2.2.1). The proposed EA optimises the same four objectives as the MOC EA (reviewed in Subsection 2.3.3), with the main difference that the proposed EA is based on lexicographic optimisation, whilst MOC is based on Pareto dominance. Section 3.1 first discusses the motivation for designing the EA based on lexicographic optimisation, and then describes the lexicographic optimisation component of the EA, based on lexicographic tournament selection. Other EA components like population initialisation, crossover and mutation operators are not discussed because they are the same operators used by MOC [Dandl et al. 2020].

The second contribution of this paper is to extend both the proposed lexicographic optimisation-based EA and MOC to make them more resilient to violations of monotonicity constraints, as a form of further improving a counterfactual's usefulness and its acceptance by users. Section 3.2 first describes the concept of monotonicity constraints and the motivation for using this concept for improving the resilience of counterfactuals, and then it describes precisely how monotonicity constraints are incorporated into the proposed lexicographic optimisation-based EA and MOC.

Note that, in all the EAs proposed in this paper (both the lexicographic optimised-based EA and the MOC extension), an individual represents a counterfactual, using the individual representation discussed in Section 2.3 (which is also the representation used by the original MOC).

### 3.1 A Novel Lexicographic Optimisation-based EA for Counterfactual Generation

The motivation for designing this EA based on the lexicographic optimisation approach, rather than the more popular Pareto dominance approach, involves three factors. The first two factors were broadly discussed in Subsection 2.2.2, but they are discussed next in the specific context of EAs for counterfactual generation, as follows.

First, although the Pareto dominance-based MOC EA (described in Subsection 2.3.3) returns a diverse set of nondominated solutions to the user, many of these returned solutions tend to be invalid counterfactuals, as shown by some experiments reported in Section 5.1. This drawback is due to the fact that the Pareto approach treats the validity objective on an equal bearing with other objectives, not recognising that, among the four objectives optimised by MOC, validity is clearly the most important objective. Hence, in a Pareto dominance-based EA like MOC, a solution with a relatively bad value of validity but relatively good values of the other three objectives can be nondominated and returned to the user, which is clearly undesirable. Validity is an essential property of a counterfactual, and invalid returned solutions, which do not satisfy the definition of a counterfactual, are in principle useless to users. The lexicographic optimisation approach naturally avoids this problem, by simply assigning to the validity objective the highest priority in the priority ordering of objectives.

---

**Procedure** Lex-Tournam-Select($n$, $k$, **O**, $\theta$, **F**)
$\mathbf{V}_{\text{FINAL}} \leftarrow \{\}$                                              // Returned set of victors, initially empty
$numV \leftarrow 0$                                                     // Number of victors found

**while** $|\mathbf{V}_{\text{FINAL}}| < n$ **do**                          // Repeat tournament rounds until $n$ victors are selected
        **P** $\leftarrow$ *sample*(**F**, $k$)                    // Randomly sample $k$ participants from the population

    *victorSelected* $\leftarrow$ FALSE
    **while not**(*victorSelected*) **do**
        **for each** $o \in \mathbf{O}$ **do**            // For each objective in the ordered list of objectives
            **P** $\leftarrow$ *sortByAsc*(**P**, $o$)        // Sort participants by current objective's fitness
            $\mathbf{p}_* \leftarrow \mathbf{P}^{[1]}$
            $\mathbf{V}_{\text{ROUND}}^{[1]} \leftarrow \mathbf{p}_*$                   // Store fittest participant as first potential victor
            $i \leftarrow 2$
            **while** $|\mathbf{p}_*^{[o]} - \mathbf{p}^{[i][o]}| \leq \theta$ **and** $i \leq |\mathbf{P}|$ **do**
                $\mathbf{V}_{\text{ROUND}}^{[i]} \leftarrow \mathbf{P}^{[i]}$            // Retain participants within tolerance threshold
                $i \leftarrow i + 1$

            **if** $|\mathbf{V}_{\text{ROUND}}| = 1$ **then**        // If a single victor has emerged
                $numV \leftarrow numV + 1$
                $\mathbf{V}_{\text{FINAL}}^{[v]} \leftarrow \mathbf{p}_*$
                *victorSelected* $\leftarrow$ TRUE
                BREAK                        // End the tournament round
            **else**
                **P** $\leftarrow \mathbf{V}_{\text{ROUND}}$             // Otherwise set as next group of participants
        **if not**(*victorSelected*) **then**       // If no victor after all objectives' comparisons
            **if** $\theta = 0$ **then**
                BREAK                       // End the tournament round if tolerance threshold is 0
            **else**
                $\theta \leftarrow 0$                       // Otherwise set tolerance threshold to 0
    **if not**(*victorSelected*) **then**       // If still no victor at the end of the tournament round
        $numV \leftarrow numV + 1$
        $\mathbf{V}_{\text{FINAL}}^{[i]} \leftarrow$ *sample*(**P**, 1)       // A victor is randomly sampled from participants
**Return** $\mathbf{V}_{\text{FINAL}}$

**Algorithm 1:** Lexicographic optimisation-based tournament selection (performing one or multiple tournament rounds)

Second, although the large number of nondominated solutions typically returned by a Pareto dominance-based EA gives the user flexibility of choice, this also presents the user with the non-trivial problem of selecting the single "best" solution for practical implementation in some real-world application [Freitas 2024]. There are automated approaches to choose the "best" out of many nondominated solutions [Petchrompo et al. 2022],[Li et al. 2020], but in general such approaches are not guaranteed to return the best solution based on a user's preferences or specific requirements of the target real-world application.

As a third factor, in general the Pareto approach incurs substantially greater computational cost during the execution of the EA than the lexicographic approach, as each population of candidate solutions must be assessed for Pareto dominance, in addition to recalculating their crowding distances [Deb et al. 2002].

The proposed EA uses the principle of lexicographic multi-objective optimisation to optimise the four objectives defined in Subsection 2.3.3. This is achieved by a procedure that performs one or more rounds of tournament selection based on lexicographic optimisation, as described in Algorithm 1.

Algorithm 1 takes as input: (1) the number $n$ of tournament rounds and also the number of victors (one for each tournament round) to be returned by this procedure; (2) the tournament size $k$ for the number of participants in each tournament round; (3) a user-defined priority ordering (a permutation) of the list of four objectives **O**; (4) a tolerance threshold $\theta$ used for comparing each objective's values between candidate solutions (to determine if there is a substantial difference between those two values); and (5) the set of fitness values **F**, one fitness value for each individual in the population.

Note that Algorithm 1 is flexible enough to be used for two different purposes within the EA, as follows. First, Algorithm 1 is executed once per generation in order to select all parents for producing offspring (via crossover and mutation). In this case, $n$ is set to the population size, which is also the number of parents which will be used to produce offspring at each generation; and $k$ is a user-defined tournament size parameter. Second, Algorithm 1 is executed at the end of the evolutionary search, in order to select the best individual of the last generation's population, which is the solution returned by the EA. In this case, Algorithm 1 is parameterised with $n = 1$ and $k$ equal to the number of individuals in the last population after removing duplicate individuals.

Regarding the priority order of objectives, we explored two different orderings, but in both orderings the validity objective was set as the highest priority objective, since validity is an essential property of a counterfactual, as mentioned earlier. The precise objective priority orderings and other hyperparameter settings for the proposed EA will be specified in Section 4 (Experimental Setup); and the precise settings of the input parameters in Algorithm 1 will be given in Subsection 4.4.

Regarding implementation issues, the proposed lexicographic optimisation-based EA was implemented by replacing the Pareto dominance-based selection procedure of MOC (Subsection 2.3.3) by the lexicographic tournament selection procedure shown in Algorithm 1; using the Evolutionary Computation in R (ECR) package – the library documentation for ECR is available at: https://cran.r-project.org/web/packages/ecr/ecr.pdf. As much as possible, all the remaining MOC code was re-used, so that the comparison of results between the proposed lexicographic optimisation-based EA and the Pareto dominance-based MOC EA can be as controlled and as fair as possible – we emphasise that both EAs are optimising the same four objectives. Hence, to maintain diverse populations of candidate solutions, the proposed EA used the same survival strategy as MOC [Dandl et al. 2020], which is a function for selecting diverse individuals based on their crowding distances.

## 3.2 Extending the Validity Objective with a Measure of Resilience to Violations of Monotonicity Constraints

In this section, we first describe the motivation to extend a counterfactual-generator EA with an additional desirable counterfactual property: resilience to violations of monotonicity constraints (defined below); and then we describe how this extension is incorporated into one of the four objectives optimised by the proposed lexicographic optimisation-based EA and by the Pareto dominance-based MOC EA – namely, the crucial objective of validity.

To introduce this notion of resilience, let us revisit the counterfactual-explanation example given in Section 2.1: "If your salary was £50,000 (rather than your current £35,000), your loan would be approved". Note that this counterfactual refers to a *precise target value* of salary, £50,000. However, attaining a precise salary value is usually impractical for users in the real world. E.g., if the user secured a new job with a salary of £55,000, it would be irrational for that user to request a reduction of their salary to £50,000 just to match the counterfactual's recommendation of £50,000; users would naturally expect that the higher the salary, the higher the probability of loan approval – an example of a "monotonicity constraint". In the classification task, such constraints in general assert that if the value of a feature monotonically increases or decreases, then the probability of the positive class should consistently increase or decrease [Sill 1997],[Cano et al. 2019]. However, due to the nonlinearity of black-box predictive models, this monotonicity constraint cannot be guaranteed. Importantly, such monotonicity constraints often align with human intuition; and intuitively a user would expect the occurrence of such monotonicity constraints when following the guidance of a counterfactual explanation. Therefore, if the user managed to increase their salary to a value larger than the target value suggested by the counterfactual, and still received a negative output (e.g. "denied loan") when reassessed by the black-box model at a later date, this would be very frustrating for the user and damaging to the reputation of the financial institution offering loan services.

Hence, an effective counterfactual explanation must not only provide a succinct path of recourse to the user, but it must also be resilient in maintaining its validity when subjected to perturbations that arise from intuitive monotonicity-constraint assumptions. Therefore, in this work we propose to use this notion of resilience to emulate monotonicity constraints in upholding expected real-world patterns, aiming to generate counterfactuals that are better trusted by users and field experts [Pazzani et al. 2021] without enforcing monotonicity constraints in the black-box model itself – which would potentially reduce its predictive accuracy [Cano et al. 2019]. This notion of resilience is also useful for improving the robustness of generated counterfactuals, which is important not only for accuracy reasons, but also for ethical and epistemic reasons [Hancox-li 2020].

Hence, we propose a formalisation of a counterfactual's *resilience to violations of monotonicity constraints (resilience* for short), a novel property of counterfactuals, and we use this new property to create two novel versions of EAs for counterfactual generation, namely an extended version of the lexicographic optimisation-based EA described in Section 3.1 and an extended version of the Pareto dominance-based MOC EA described in Subsection 2.3.3. Note that a counterfactual *must be valid* in order to have the property of resilience and therefore be eligible for resilience testing.

Counterfactual resilience measures the ability of a counterfactual to maintain its validity (predicting its target class) as the values of the features suggested for change are increased or decreased to a given upper or lower bound, such as the maximum or minimum value of the feature across all instances in the observed training set. This definition pertains solely to features with *numeric* values. This definition also pertains to *univariate* resilience, in that incremental and decremental mutations on feature values are carried out individually, i.e. when one feature is being mutated, the values of all other features of the counterfactual are preserved. If a counterfactual $x_{cf}$ contains a numeric feature's value that is different from that feature's value in the original data point of interest $x_{pt}$, then its resilience can be measured. Let $x_{cf}^{[i]}$ and $x_{pt}^{[i]}$ denote the value of the $i$-th numeric feature in $x_{cf}$ and $x_{pt}$. To measure the resilience of a counterfactual $x_{cf}$, for each $i$-th numeric feature with a suggested change in $x_{cf}$ (i.e., $x_{cf}^{[i]} \neq x_{pt}^{[i]}$), that feature value $x_{cf}^{[i]}$ is incrementally increased or decreased in normalised steps to the upper or lower bound of that $i$-th feature in the training set, thus mutating the counterfactual. The value of each mutation step is given by:

$$step = \begin{cases} ((U^{[i]} - x_{cf}^{[i]}) \div 10), & if \ x_{cf}^{[i]} > x_{pt}^{[i]} \\ ((L^{[i]} - x_{cf}^{[i]}) \div 10), & otherwise \end{cases}$$

where $U^{[i]}$ and $L^{[i]}$ are the upper bound and lower bound (respectively) of the $i$-th changed numeric feature in $x_{cf}$. This incorporates the polarity of the step value. If the feature is integer-valued, then the step value is rounded to the nearest integer, unless this rounding would lead to 0, in which case the step is set to 1 if $x_{cf}^{[i]} > x_{pt}^{[i]}$ or 0 otherwise.

After the step is calculated, the maximum number of mutative steps is given by:

$$step_{max} = \begin{cases} |\lfloor (U^{[i]} - x_{cf}^{[i]}) \div step \rfloor|, & if \ x_{cf}^{[i]} > x_{pt}^{[i]} \\ |\lfloor (L^{[i]} - x_{cf}^{[i]}) \div step \rfloor|, & otherwise \end{cases}$$

where $\lfloor x \rfloor$ is the "floor" of x (the largest integer that is smaller than or equal to x); and | x | is the absolute value of x.

The full calculation of a counterfactual's resilience scores is shown in Algorithm 2. At each step, the mutated counterfactual $\delta x_{cf}$ is reclassified by the black-box model. If $\hat{f}(\delta x_{cf}) = \hat{f}(x_{cf})$, then $\delta x_{cf}$ is still valid and the next step of mutation is carried out. If $\hat{f}(\delta x_{cf}) \neq \hat{f}(x_{cf})$, then $\delta x_{cf}$ is no longer valid and the mutation steps halt. If a counterfactual's feature value is already at or beyond the upper or lower bound for that feature (observed in the training set), no mutative step can be applied, and this feature is considered fully resilient for the given counterfactual.

---

**procedure** GetResilience($x_{cf}$, $x_{pt}$, $\hat{f}$, **U**, **L**)
  **resScores** ← **0**
  **for each** changed numeric feature $i$ such that $x_{cf}^{[i]} \neq x_{pt}^{[i]}$ **do**

    **if** $x_{cf}^{[i]} \geq U^{[i]}$ **or** $x_{cf}^{[i]} \leq L^{[i]}$ **then**      // If at or beyond the upper/lower bound,
      **rescores**[i] ← 1            // the feature is fully resilient for $\delta x_{cf}$
    **else**
      *bound* ← NULL
      *val* ← NULL
      **if** $x_{cf}^{[i]} > x_{pt}^{[i]}$ **then**      // If counterfactual feature increased i-th feature value
        *bound* ← $U^{[i]}$          // Set bound to upper bound to $i$-th feature
        *val* ← 1                // Step value is incremental
      **else**                  // Otherwise counterfactual decreased i-th feature value
        *bound* ← $L^{[i]}$          // Set bound to lower bound to $i$-th feature
        *val* ← −1               // Step value is decremental
      *step* ← (*bound* − $x_{cf}^{[i]}$) ÷ 10

      **if** *isInteger*($x_{cf}^{[i]}$) **then**
        *step* ← *round*(*step*)
        **if** *step* = 0 **then**
          *step* ← *val*          // Set step to 1/-1 if rounding gives 0

      *steps$_{max}$* ← $|\lfloor$(*bound* − $x_{cf}^{[i]}$) ÷ *step*$\rfloor|$

      $\delta x_{cf}^{[i]}$ ← $\delta x_{cf}^{[i]}$ + *step*      // Mutate $i$-th feature by step value
      **if** $\hat{f}(\delta x_{cf}) \neq \hat{f}(x_{cf})$ **then**    // If mutated counterfactual $\delta x_{cf}$ becomes invalid
        BREAK              // Resilience-testing limit reached: Terminate
      *steps$_+$* ← *steps$_+$* + 1

    **resScores**[i] ← *steps$_+$* ÷ *steps$_{max}$*    // Store resilience score for $i$-th feature
  **return resScores**

---

**Algorithm 2:** Procedure for generating an array of feature resilience scores for a given counterfactual

The resilience of any changed numeric feature $x_{cf}^{[i]}$ is given by the number of *successful* mutative steps to a given bound without producing an invalid counterfactual divided by the *maximum* number of mutative steps. The resilience function ($R$) returns a set of normalised resilience scores $r^{[1]},\ldots r^{[p]}$, where $p$ is the number of mutated numeric features in $x_{cf}$ and each feature's resilience score $r^{[i]}$ (for $i$ in $[1,\ldots,p]$) is a value between 0 and 1, where 0 means the $i$-th feature in $x_{cf}$ has *no resilience* and 1 means that feature is *fully resilient*.

$$R : (x_{pt}, x_{cf}, \hat{f}, U, L) \to (r^{[1]},\ldots,r^{[p]})$$

The overall resilience score for the counterfactual $x_{cf}$ is given by the mean average of its feature resilience scores over all $p$ numeric features whose values in the counterfactual are different than those features' values in the point of interest $x_{pt}$:

$$\overline{R} = \frac{1}{p} \sum_{i=1}^{p} r^{[i]}$$

Given that resilience is the ability for a counterfactual to maintain its validity when subjected to continuous unidirectional mutations, the resilience score is integrated into the objective of validity, which is one of the objectives optimised by the multi-objective EAs in this work. Hence, the new first objective ($o_1$), validity integrating resilience, is defined as follows:

$$o_1(\hat{f}_p(\delta x_{pt}), Y') = \begin{cases} 0 - \overline{R}(\delta x_{pt}), & \text{if } \hat{f}_p(\delta x_{pt}) \in Y' \\ \inf_{y' \in Y'} |\hat{f}_p(\delta x_{pt}) - y'|, & \text{else} \end{cases}$$

Note that a candidate solution is denoted as $\delta x_{pt}$ (rather than $x_{cf}$) in this equation because it is not yet verified to be a valid counterfactual. However, in the pseudocode for calculating resilience, an individual is notated as $x_{cf}$ (and $\delta x_{cf}$ when being mutated) because it has been verified to be a valid counterfactual that is eligible for resilience testing. With integrated resilience, the range of continuous values for objective $o_1$ is [–1, 0.5]. Values between –1 and 0 indicate valid counterfactuals, where –1 indicates a fully resilient counterfactual and 0 a zero-resilient counterfactual. Thus, ideal values remain minimal, consistent with all other objectives being optimised by the EAs in this work.

**4 Experimental Setup**
The experiments tested the effectiveness of the proposed EA with lexicographic tournament selection (Section 3.1) against the original MOC EA with the Pareto-based nondominated selection (Subsection 2.3.3), in addition to testing the effectiveness of the new validity-with-resistance objective and its impact on the performance of each of the above two types of EAs.

To produce a broad understanding of each EA's performance, the experiments used five classification datasets (Section 4.1) and three machine learning algorithms that learn black-box classification models (Section 4.2). Each model was trained and tuned using the Machine Learning in R (MLR) package (https://cran.r-project.org/web/packages/ecr/ecr.pdf).

**4.1 The Five Datasets Used in the Experiments**
The five datasets have a varied number of features and instances, as shown in Table 1. These datasets are in general widely used in the literature for machine learning explainability.

**Table 1:** Main characteristics of the datasets used in the experiments

|  | Adult Income | Compas | Diabetes | FICO | German credit |
|---|---|---|---|---|---|
| # instances in original dataset | 32,561 | 7,214 | 768 | 10,459 | 1,000 |
| # instances with no missing value | 30,162 | 6,907 | 392 | 9,871 | 522 |
| # training instances | 29,662 | 6,407 | 262 | 9,371 | 348 |
| # testing instances | 500 | 500 | 130 | 500 | 174 |
| # classes | 2 | 2 | 2 | 2 | 2 |
| # features | 14 | 10 | 8 | 23 | 9 |
| # non-actionable features | 7 | 4 | 2 | 1 | 2 |

These datasets represent three broad types of application domains: finance, law and medicine; and a brief summary of their main purpose is as follows.
- Adult Income Census dataset (Adult)– used for predicting individual income based on various indicators, including financial, occupational and educational indicators. (Available from: https://archive.ics.uci.edu/dataset/2/adult)
- COMPAS Recidivism Risk Score (Compas) – used for assessing the probability of repeat offences based on criminal history and for identifying potential racial bias. (https://github.com/propublica/compas-analysis/)
- Pima Indians Diabetes (Diabetes) – used for diabetes diagnosis based on biomedical indicators. (https://www.kaggle.com/datasets/uciml/pima-indians-diabetes-database)

- FICO Explainable Machine Learning (FICO) – used for credit risk assessment based on financial indicators. (https://community.fico.com/s/explainable-machine-learning-challenge)
- German Credit Risk (German credit) – used for credit risk assessment based on financial and occupational indicators. (https://github.com.dandls/moc/tree/master/credit_example)

Since the algorithms in the MLR library cannot handle missing values, all instances that contain missing values were removed from the dataset in a data preprocessing step, which is the same approach used in [Dandl et al. 2020]. Each dataset was split into a training set and a test set; the former for training the model and the latter for providing points of interest (instances) for which counterfactuals are generated by the EA runs. For the three largest datasets (Adult, Compas, FICO), the test set was capped at 500 instances. For the two smallest datasets (Diabetes, German credit), the number of instances in the test set was capped at one third of the full dataset without missing values.

Note that non-actionable features are features that users cannot actionably change, e.g. race. The non-actionable features for these datasets were identified based on those defined as non-actionable in Section 6.1 of [Guidotti 2024]. The EAs evaluated in this work handle non-actionable features by taking a set of feature names that are designated as non-actionable by the user, and then those features are not mutated throughout the EA run. Therefore, the counterfactuals returned by all EA runs in our experiments do not contain any nonactionable feature. The list of non-actionable features for each dataset is provided in the Appendix.

### 4.2 The Three Classification Algorithms Used for Learning Black-Box Models

In the experiments, the EAs generated counterfactuals for black-box classification models learned by three types of classification algorithms: Neural Network (NNet), Random Forest (RF) and Support Vector Machine (SVM). Each model was trained and tuned using the MLR library. Although maximising the black-box models' predictive accuracy is not a goal of this work (our goal is rather to maximise the quality of the counterfactuals generated by the EAs), we performed some basic hyperparameter tuning for these algorithms, to minimise the risk that a learned black-box model would have a relatively low predictive accuracy due to a poor hyperparameter setting.

Hyperparameter tuning was performed by applying random search (a simple and popular method for hyperparameter tuning [Bergstra & Bengio 2012]) to the training set, in order to choose the best configuration of each algorithm among the range of candidate hyperparameter settings shown in Table 2. As shown in this table, the tuned hyperparameters were: (a) for NNet: *size* is the number of hidden layers in the network, and *decay* is the rate of learning rate reduction over time; (b) for RF: *ntree* is the number of decision trees in the forest, and *mtry* is the number of randomly sampled features that are evaluated at each tree node; (c) for SVM: *cost* indicates the trade-off between maximising the separation between classes and minimising the classification error rate.

**Table 2:** The classification algorithms' hyperparameters tuned in the experiments

|  | Neural Net | | Random Forest | | SVM |
|---|---|---|---|---|---|
| Hyperparameter | size | decay | ntree | mtry | cost |
| Range of values used for tuning | [1 : 5] | [0.1 : 0.9] | [50 : 500] | [1 : #features] | [0.01 : 1] |

### 4.3 Sampling Points of Interest (Test Instances) To Be Explained by Counterfactuals

For each of the 15 combinations of 5 datasets times 3 classification algorithms, the experiments proceeded as follows. First, the dataset was divided into training and testing sets, and then a black-box classification model was trained, as described in Sections 4.1 and 4.2. Then, instances were randomly sampled, without replacement, from the testing set, to get the points of interest for which counterfactuals would be generated.

Each sampled testing instance is classified by the black-box model. If the model predicts the positive class for that instance, it is discarded; whilst if the model predicts the negative class, the sampled instance is used as a point of interest for counterfactual generation. We focus on using only testing instances with originally predicted negative class as points of interest because this is the most useful scenario for counterfactual generation in practice, assuming that the positive class is the target class of interest. I.e., in general, it is more interesting or useful to find a counterfactual that suggests how an individual's characteristics should be changed in order to switch the black-box prediction from negative to positive class, rather than the other way around.

### 4.4 The Configurations of the EAs Used in the Experiments

The experiments compare two versions of the proposed lexicographic optimisation-based EAs, hereafter referred to as Lex-EAs, against the original MOC EA [Dandl et al. 2020], which is a Pareto dominance-based EA hereafter referred to as Par-EA. Both the Lex-EAs and Par-EA use the same individual representation, the same genetic operators, and the same hyperparameter settings for the parameters they have in common, in order to make their comparison as controlled and as fair as possible.

Par-EA optimised the four objectives discussed in Section 2.3.3. The two versions of Lex-EA also optimise those four objectives, they differ only in the priority order of objectives to be optimised. In both Lex-EA versions, the highest-priority objective is validity or the proposed variation of validity with integrated resilience (Section 3.2), since validity is intuitively the most important property of a counterfactual. Actually, a candidate solution should not be considered a counterfactual if validity is not satisfied. In addition, in both EA versions the lowest-priority objective is plausibility.

Sparsity and distance to point of interest are intuitively more important than plausibility because the former two objectives directly measure how many of their characteristics (features) an individual (instance) would have to change, and what is the required magnitude of those changes, in order for an individual to be re-classified as a positive instance by the black-box model. By contrast, plausibility measures to what extent the feature values in the counterfactual are consistent with feature values in the training set as a whole, rather than focusing specifically on the point of interest being explained. In addition, broadly speaking, sparsity and distance to point of interest are more often used as objectives (or are given a greater weight in the weight-sum approach) than plausibility, in the counterfactual-generator EA literature. For example, as discussed in Section 2.3, CERTIFAI is a single-objective EA that optimises distance to point of interest only; whilst the GeCo EA optimises validity, sparsity and distance to point of interest – i.e. none of these two EAs optimises plausibility. However, the choice between prioritising sparsity over distance to point of interest or vice-versa is not clear. For example, CERTIFAI does not optimise sparsity, but GeCo prioritizes sparsity over distance to point of interest.

Hence, we explored two versions of Lex-EA, varying the priority ordering between sparsity and distance to point of interest, but always fixing validity (with or without integrated resilience) and plausibility as the highest-priority and the lowest-priority objectives, respectively. The two EA versions, hereafter denoted Lex-EA-1 and Lex-EA-2, use the following objective priority orderings:

Lex-EA-1: validity, distance to point of interest, sparsity, plausibility
Lex-EA-2: validity, sparsity, distance to point of interest, plausibility

Each of the three base versions of an EA (Par-EA, Lex-EA-1, Lex-EA-2) was tested with two versions of the validity objective: with and without integrated resilience (see Section 3.2), forming in total six EA versions.

The lexicographic selection procedure (Algorithm 1) in both Lex-EAs used the following input parameter values. First, recall that as discussed in Section 3.1, Algorithm 1 is used for two purposes in the EA. When Algorithm 1 is used to select all parents to produce offspring in each generation, the algorithm is parameterised with $n = 20$ tournament rounds to return 20 victors (as the population size given by MOC's original configuration file is 20) and $k = 2$ for the tournament size, such that a full population of 20 parents (which will produce offspring) is returned from each call of Algorithm 1. When Algorithm 1 is used to select the best individual at the end of the EA's search (i.e. the individual to be returned by the EA as the best found solution), the algorithm is parameterised with $n = 1$ tournament round to return one victor, and $k$ equals the number of individuals in the final generation after removing duplicate individuals. In both these uses of Algorithm 1, the tolerance threshold $\theta$ was set to 0.01, a common value in the lexicographic optimisation literature.

With the exception of the EA parameters specific to the Lex-EA versions (which were set as mentioned above), all other EA parameters (such as crossover and mutation probabilities, population size, etc.) are shared by all 6 EA versions, and all those EA versions were run with the same setting for those parameters, which are the settings used by the original Par-EA [Dandl et al. 2020]. These settings were retrieved from the configuration file available in that work's repository, at the GitHub site: https://github.com/dandls/moc/blob/master/saved_objects/best_configs.rds. In that work, these parameter settings were tuned by using iterated F-racing (irace) [Lopez-Ibanez et al. 2021], a well-known tool for parameter optimisation of meta-heuristics like EAs. Note that these parameter settings were tuned for Par-EA only, and we have made no attempt to tune these parameter settings for Lex-EA-1 and Lex-EA-2. Hence, from the perspective of parameter tuning, the results of our experiments might be slightly biased in favour of Par-EA, but we have chosen to use the same parameter settings for both Par-EA and the two Lex-EAs in order to make the comparison between all EA versions as controlled and as fair as possible. A run of Par-EA stops after a set number of generations or until a convergence condition is met. In either case, the number of generations actually executed for each Par-EA run is also used as the number of generations executed by their counterpart Lex-EA-1 and Lex-EA-2 runs, so that all 6 EA versions use the same "computational budget" (the same populations size and the same number of generations), again for a controlled and fair comparison.

Comparisons between Par-EA's results and the two Lex-EA's results, as well as resilience testing, are only possible if all three EAs return counterfactuals, i.e. none of them return "null". Resilience testing is only possible if each EA returns at least one counterfactual that is valid and has changed numeric features from the original data point of interest. For each combination of a dataset and a type of black-box model, points of interest are sampled from the test set (see Section 4.3) and used for producing counterfactuals by each EA until a maximum of 50 points of interest have produced counterfactuals and have been resilience-tested, or until the test set is empty. The experimental results reported in the next Section are an average over the experiments with the about 50 points of interest used for each combination of a dataset and a black-box model.

## 5 Computational Results and Discussion

This Section is divided into three parts. Subsection 5.1 reports the results regarding the numbers of counterfactuals returned by each EA and the percentages of those counterfactuals that are valid, i.e., predicting the target (positive) class. Subsection 5.2 reports the values of the four objectives optimised by the EAs, in the counterfactuals returned by those EAs. This will allow us to determine which EA achieve the best value *for each of those four objectives*. Subsection 5.3 reports the results of comparing the lexicographic optimisation-based EAs (Lex-EAs) with the Pareto dominance-based EAs (Par-EAs) considering the quality of their returned counterfactual solutions from a holistic perspective, i.e. considering *the four objectives as a whole*. Note that each table of results reported in this Section shows results for 6 EA versions: 3 EAs, namely Par-EA, Lex-EA-1 and Lex-EA-2, each with two versions, namely with or without integrating resilience into the validity objective.

### 5.1 Results Regarding the Percentages of Valid Counterfactuals Returned by Each EA

Table 3 reports the number of counterfactuals returned by each EA version as well as (between brackets) the percentage of valid counterfactuals among those returned counterfactuals. This table has one row of results for each of the 30 combinations of 3 types of black-box models times 5 datasets times 2 EA versions depending on whether or not a measure of resilience to violations of monotonicity constraints is integrated into the validity objective. The result in each table cell is the average result over all counterfactuals returned by all runs of the EA version associated with that cell. The last 3 rows of the table show three types of average results: average over the 15 experimental settings for each EA version that does not use resilience, average over the 15 settings for each EA version that uses resilience, and the overall average across all 30 experiments in the table. This table structure is also used for other tables of results.

Note that Par-EA obviously returns many more counterfactuals than the Lex-EAs by design, since Par-EA returns the best found Pareto front in each run, whilst the Lex-EAs return a single counterfactual per run. Overall there is very little difference between the number of counterfactuals returned by Lex-EA-1 and Lex-EA-2, as shown in Table 3. Hence, the most interesting patterns in this table refer to the percentage of valid counterfactuals achieved by each EA, as follows.

First, as shown in the last row of Table 3 (with average results over all 30 experiments), the percentage of valid counterfactuals returned by Lex-EA-1 and Lex-EA-2 is about 94%, whilst the percentage of valid counterfactuals returned by Par-EA is much smaller, about 45%. These results are qualitatively similar in the scenarios with or without resilience integrated into the validity objective, as shown in the last-but-one and last-but-two rows in Table 3.

Those two rows also show that using resilience had a minor positive effect on the percentage of valid counterfactuals returned by Par-EA, with an increase of 1%; whilst using resilience had a substantial positive effective on the percentage of valid counterfactuals returned by Lex-EA-1 and Lex-EA-2, with increases of about 9% and 8%, respectively.

In summary, the two Lex-EAs achieved a much higher percentage of valid counterfactuals than Par-EA; and using resilience integrated into the validity objective was substantially beneficial to the two Lex-EAs, whilst having almost no effect for Par-EA. This can be explained by the fact that the objective integrating resilience and validity is given the highest priority (among all objectives) in the in the Lex-EAs, but that objective is not given any special priority in Par-EA.

**Table 3:** Number of counterfactuals (and % of valid counterfactuals) returned by each EA, for each combination of a type of black-box model, a dataset, and whether or not the EA uses resilience integrated into the validity objective

| Type of model | Dataset | EA with resilience? | EA version (Pareto or Lexicographic (v1 or v2)) | | |
|---|---|---|---|---|---|
| | | | Par-EA | Lex-EA-1 | Lex-EA-2 |
| Neural network | Adult | no | 2,781 (40.4%) | 55 (96.4%) | 55 (92.7%) |
| | | yes | 3,248 (32.6%) | 50 (100%) | 50 (100%) |
| | Compas | no | 1,195 (33.3%) | 51 (98%) | 51 (98%) |
| | | yes | 1,325 (38.2%) | 50 (100%) | 50 (100%) |
| | Diabetes | no | 4,755 (61.6%) | 51 (100%) | 51 (98%) |
| | | yes | 5,095 (59.6%) | 51 (100%) | 51 (98%) |
| | FICO | no | 1,300 (47.5%) | 52 (96.2%) | 52 (98.1%) |
| | | yes | 1,368 (47.7%) | 50 (100%) | 50 (100%) |
| | German credit | no | 2,355 (63.1%) | 50 (100%) | 50 (100%) |
| | | yes | 3,203 (67%) | 50 (100%) | 50 (100%) |
| Random Forest | Adult | no | 3,734 (58.3%) | 58 (91.4%) | 58 (91.4%) |
| | | yes | 3,367 (49.4%) | 50 (100%) | 50 (100%) |
| | Compas | no | 986 (38.3%) | 51 (100%) | 51 (98%) |
| | | yes | 1,301 (46.7%) | 50 (100%) | 50 (100%) |
| | Diabetes | no | 4,752 (51.5%) | 53 (84.9%) | 53 (92.5%) |
| | | yes | 5,882 (48.8%) | 50 (100%) | 50 (100%) |
| | FICO | no | 2,546 (33.7%) | 84 (77.4%) | 84 (73.8%) |
| | | yes | 1,882 (38.4%) | 50 (100%) | 50 (100%) |
| | German credit | no | 3,523 (63%) | 55 (92.7%) | 55 (90.9%) |
| | | yes | 3,364 (65.3%) | 50 (100%) | 50 (100%) |
| SVM | Adult | no | 3,327 (38.6%) | 54 (92.6%) | 54 (96.3%) |
| | | yes | 4,189 (51.8%) | 51 (98%) | 51 (98%) |
| | Compas | no | 2,332 (30.4%) | 51 (100%) | 51 (98%) |
| | | yes | 3,014 (32.4%) | 50 (100%) | 50 (100% |
| | Diabetes | no | 5,350 (41.3%) | 50 (84%) | 50 (84%) |
| | | yes | 6,716 (37.3%) | 50 (98%) | 50 (98%) |
| | FICO | no | 18,819 (17%) | 196 (42.4%) | 196 (43.3%) |
| | | yes | 10,137 (21.6%) | 76 (76.3%) | 76 (69.7%) |
| | German credit | no | 5,192 (52%) | 79 (83.5%) | 79 (86.1%) |
| | | yes | 4,517 (49.26%) | 50 (100%) | 50 (100%) |
| Mean | | no | 4,196.5 (44.7%) | 66 (89.3%) | 66 (89.4%) |
| | | yes | 3,907.2 (45.7%) | 51.9 (98.2%) | 51.9 (97.6%) |
| | | overall | 4,051.8 (45.2%) | 58.9 (93.7%) | 58.9 (93.5%) |

## 5.2 Results Regarding the Optimised Values of Each of the Four Objectives for Each EA

Table 4 reports the average values of Objective 1, involving the property of validity, among the solutions returned by each EA. There are two versions of this objective, as follows. In the EA versions where this objective is validity without resilience, the objective's value for a solution is 0 if it is a valid counterfactual; otherwise this objective value indicates the distance between the predicted probability of the positive class by the model ($\hat{f}_p$) and the target range Y' = [0.5, 1] (denoted Dist.target.class in Table 4). In the EA versions where this objective is validity with integrated resilience (called Resil.validity in Table 4), objective values between -1 and 0 indicate a valid counterfactual, where -1 is fully resilient and 0 is zero-resilient; otherwise the objective value indicates the value of Dist.target.class, as defined above.

**Table 4:** Average values of Objective 1 achieved by each EA, for each combination of a type of black-box model, a dataset, and a specification of Objective 1: Resil.validity or Dist.target.class, depending on whether or not (respectively) the EA uses validity with resilience

| Type of model | Dataset | EA with resilience? | Objective being optimised | EA version | | |
|---|---|---|---|---|---|---|
| | | | | Par-EA | Lex-EA-1 | Lex-EA-2 |
| Neural network | Adult | no | Dist.target.class | 0.139 | 0.0 | 0.0 |
| | | yes | Resil.validity | -0.105 | -1.0 | -1.0 |
| | Compas | no | Dist.target.class | 0.244 | 0.0 | 0.0 |
| | | yes | Resil.validity | -0.134 | -1.0 | -1.0 |
| | Diabetes | no | Dist.target.class | 0.112 | 0.0 | 0.005 |
| | | yes | Resil.validity | -0.406 | -1.0 | -0.976 |
| | FICO | no | Dist.target.class | 0.098 | 0.0 | 0.0 |
| | | yes | Resil.validity | -0.356 | -1.0 | -1.0 |
| | German credit | no | Dist.target.class | 0.074 | 0.0 | 0.0 |
| | | yes | Resil.validity | -0.483 | -1.0 | -1.0 |
| Random Forest | Adult | no | Dist.target.class | 0.141 | 0.001 | 0.001 |
| | | yes | Resil.validity | -0.252 | -1.0 | -1.0 |
| | Compas | no | Dist.target.class | 0.216 | 0.0 | 0.0 |
| | | yes | Resil.validity | -0.282 | -1.0 | -1.0 |
| | Diabetes | no | Dist.target.class | 0.06 | 0.001 | 0.0 |
| | | yes | Resil.validity | -0.354 | -1.0 | -0.99 |
| | FICO | no | Dist.target.class | 0.068 | 0.001 | 0.001 |
| | | yes | Resil.validity | -0.287 | -0.988 | -0.996 |
| | German credit | no | Dist.target.class | 0.044 | 0.0 | 0.0 |
| | | yes | Resil.validity | -0.362 | -1.0 | -1.0 |
| SVM | Adult | no | Dist.target | 0.182 | 0.007 | 0.007 |
| | | yes | Resil.validity | -0.307 | -0.968 | -0.965 |
| | Compas | no | Dist.target.class | 0.269 | 0.0 | 0.0 |
| | | yes | Resil.validity | -0.053 | -1.0 | -1.0 |
| | Diabetes | no | Dist.target.class | 0.138 | 0.001 | 0.001 |
| | | yes | Resil.validity | -0.210 | -0.978 | -0.958 |
| | FICO | no | Dist.target.class | 0.128 | 0.053 | 0.050 |
| | | yes | Resil.validity | -0.01 | -0.568 | -0.512 |
| | German credit | no | Dist.target.class | 0.049 | 0.001 | 0.0 |
| | | yes | Resil.validity | -0.391 | -1.0 | -1.0 |
| Mean | | no | Dist.target.class | 0.131 | 0.004 | 0.004 |
| | | yes | Resil.validity | -0.266 | -0.967 | -0.96 |

As shown in the last-but-one row in Table 4, with the average results for the EAs optimising validity without resilience, both Lex-EAs achieved on average a validity (Dist.target.class) value of 0.004, substantially better than the average value of 0.131 achieved by Par-EA. In the scenario where the EAs optimise validity with resilience, as shown in the last row in Table 4, both Lex-EAs achieved on average a validity (Resil.validity) value of about -0.96, much better than the mean value of -0.266 achieved by Par-EA. Note that it is not appropriate to compute the average results over all 30 experiments in Table 4, because the objective functions being optimised are different in the two scenarios with and without resilience.

Table 5 reports the average values of Objective 2, dist.x.interest, the Gower distance between a counterfactual returned by the EA and the original point of interest (test instance being classified), for the counterfactuals returned by each EA. As shown in the last row of Table 5, overall (across all 30 experiments) the dist.x.interest value achieved by Par-EA (0.042) was somewhat smaller (better) than values achieved by Lex-EA-1 and Lex-EA-2 (0.058 and 0.061, respectively). A similar pattern was observed in the two scenarios with or without resilience, as shown in the last-but-one and last-but-two rows in Table 5. Overall, the differences between the results of all EAs are small.

In addition, the last-but-one and last-but-two rows of Table 5 show that every EA achieved very similar Objective 2 values in the two settings of using or not using resilience; i.e., using resilience had a negligible result in the EAs' ability to minimise this objective.

**Table 5:** Average values of Objective 2 (dist.x.interest) achieved by each EA, for each combination of a type of black-box model, a dataset, and whether or not the EA uses validity with resilience

| Type of model | Dataset | EA with resilience? | EA version (Pareto or Lexicographic (v1 or v2)) | | |
|---|---|---|---|---|---|
| | | | Par-EA | Lex-EA-1 | Lex-EA-2 |
| Neural network | Adult | no | 0.019 | 0.026 | 0.032 |
| | | yes | 0.021 | 0.031 | 0.033 |
| | Compas | no | 0.047 | 0.078 | 0.08 |
| | | yes | 0.049 | 0.085 | 0.085 |
| | Diabetes | no | 0.04 | 0.064 | 0.06 |
| | | yes | 0.04 | 0.057 | 0.06 |
| | FICO | no | 0.008 | 0.03 | 0.034 |
| | | yes | 0.007 | 0.021 | 0.026 |
| | German credit | no | 0.083 | 0.056 | 0.056 |
| | | yes | 0.073 | 0.055 | 0.057 |
| Random Forest | Adult | no | 0.03 | 0.038 | 0.045 |
| | | yes | 0.026 | 0.04 | 0.041 |
| | Compas | no | 0.057 | 0.077 | 0.078 |
| | | yes | 0.061 | 0.076 | 0.077 |
| | Diabetes | no | 0.029 | 0.063 | 0.063 |
| | | yes | 0.036 | 0.054 | 0.066 |
| | FICO | no | 0.034 | 0.090 | 0.101 |
| | | yes | 0.029 | 0.073 | 0.08 |
| | German credit | no | 0.062 | 0.034 | 0.039 |
| | | yes | 0.061 | 0.04 | 0.044 |
| SVM | Adult | no | 0.029 | 0.04 | 0.055 |
| | | yes | 0.026 | 0.042 | 0.052 |
| | Compas | no | 0.041 | 0.071 | 0.077 |
| | | yes | 0.044 | 0.076 | 0.076 |
| | Diabetes | no | 0.045 | 0.077 | 0.089 |
| | | yes | 0.042 | 0.079 | 0.078 |
| | FICO | no | 0.023 | 0.038 | 0.04 |
| | | yes | 0.026 | 0.044 | 0.041 |
| | German credit | no | 0.078 | 0.082 | 0.085 |
| | | yes | 0.08 | 0.092 | 0.091 |
| Mean | | no | 0.042 | 0.058 | 0.062 |
| | | yes | 0.041 | 0.058 | 0.06 |
| | | overall | 0.042 | 0.058 | 0.061 |

Table 6 reports the average values of Objective 3, sparsity, i.e., the number of features in the returned counterfactuals with a feature value different from the corresponding value in the original point of interest, for the counterfactuals returned by each EA.

As shown in the last row of Table 6, overall (across the 30 experiments) the differences between the mean sparsity values achieved by the two Lex-EAs and by Par-EA were small; the smallest (best) mean value of this objective was achieved by Lex-EA-2 (1.58), followed by Lex-EA-1 (1.73), whilst Par-EA obtained the worst mean result for this objective (1.86).

A similar pattern was observed in the two scenarios with or without resilience, as shown in the last-but-one and last-but-two rows in Table 6; with Lex-EA-2 obtaining the best mean result and Par-EA obtaining the worst mean result for the sparsity objective on both scenarios. In addition, using resilience had just a small effect on the ability of Lex-EAs and Par-EA for optimising sparsity; as each EA achieved similar results in the last-but-one and last-but-two rows of Table 6.

**Table 6:** Values of Objective 3 (sparsity) achieved by each EA, for each combination of a type of black-box model, a dataset, and whether or not the EA uses validity with resilience

| Type of model | Dataset | EA with resilience? | EA version (Pareto or Lexicographic (v1 or v2)) | | |
|---|---|---|---|---|---|
| | | | Par-EA | Lex-EA-1 | Lex-EA-2 |
| Neural network | Adult | no | 1.342 | 1.127 | 0.982 |
| | | yes | 1.454 | 1.16 | 1.0 |
| | Compas | no | 1.274 | 1.157 | 1.059 |
| | | yes | 1.275 | 1.140 | 1.08 |
| | Diabetes | no | 1.622 | 1.314 | 1.118 |
| | | yes | 1.636 | 1.157 | 1.098 |
| | FICO | no | 1.840 | 2.250 | 1.769 |
| | | yes | 1.833 | 2.06 | 1.38 |
| | German credit | no | 1.892 | 1.14 | 1.08 |
| | | yes | 1.872 | 1.16 | 1.08 |
| Random Forest | Adult | no | 1.547 | 1.293 | 1.017 |
| | | yes | 1.551 | 1.40 | 1.02 |
| | Compas | no | 1.433 | 1.059 | 1.059 |
| | | yes | 1.464 | 1.04 | 1.02 |
| | Diabetes | no | 1.537 | 1.566 | 1.245 |
| | | yes | 1.698 | 1.3 | 1.2 |
| | FICO | no | 3.091 | 5.274 | 4.524 |
| | | yes | 2.886 | 2.48 | 3.44 |
| | German credit | no | 1.721 | 1.0 | 1.036 |
| | | yes | 1.765 | 1.12 | 1.12 |
| SVM | Adult | no | 1.524 | 1.296 | 1.13 |
| | | yes | 1.544 | 1.373 | 1.078 |
| | Compas | no | 1.499 | 1.118 | 1.02 |
| | | yes | 1.605 | 1.12 | 1.0 |
| | Diabetes | no | 1.586 | 1.46 | 1.46 |
| | | yes | 1.649 | 1.580 | 1.280 |
| | FICO | no | 3.913 | 4.857 | 4.556 |
| | | yes | 4.176 | 5.184 | 4.921 |
| | German credit | no | 1.784 | 1.063 | 1.051 |
| | | yes | 1.836 | 1.58 | 1.64 |
| Mean | | no | 1.84 | 1.798 | 1.607 |
| | | yes | 1.883 | 1.657 | 1.557 |
| | | overall | 1.862 | 1.728 | 1.582 |

Table 7 reports the average values of Objective 4, plausibility, for the counterfactuals returned by each EA. As shown in the last row of Table 7, overall (across the 30 experiments) the differences between the mean plausibility values achieved by the two Lex-EAs and by Par-EA were small; the smallest (best) mean value of this objective was achieved by Par-EA (0.042), whilst the two Lex-EAs achieved slightly higher values, 0.067 and 0.069.

A very similar pattern was observed in each of the two scenarios with or without resilience, for each EA, as shown in the last-but-one and last-but-two rows in Table 7. Hence, using resilience had no effect or very little effect in the EAs' ability to optimise plausibility.

**Table 7:** Average values of Objective 4 (plausibility, measured as distance to nearest training instance) achieved by each EA, for each combination of a type of black-box model, a dataset, and whether or not the EA uses validity with resilience.

| Type of model | Dataset | EA with resilience? | EA version (Pareto or Lexicographic (v1 or v2)) | | |
|---|---|---|---|---|---|
| | | | Par-EA | Lex-EA-1 | Lex-EA-2 |
| Neural network | Adult | no | 0.019 | 0.052 | 0.057 |
| | | yes | 0.027 | 0.054 | 0.055 |
| | Compas | no | 0.02 | 0.035 | 0.033 |
| | | yes | 0.02 | 0.034 | 0.033 |
| | Diabetes | no | 0.068 | 0.1 | 0.098 |
| | | yes | 0.067 | 0.097 | 0.097 |
| | FICO | no | 0.032 | 0.054 | 0.057 |
| | | yes | 0.031 | 0.046 | 0.053 |
| | German credit | no | 0.05 | 0.088 | 0.082 |
| | | yes | 0.044 | 0.087 | 0.09 |
| Random Forest | Adult | no | 0.03 | 0.06 | 0.066 |
| | | yes | 0.03 | 0.063 | 0.065 |
| | Compas | no | 0.018 | 0.046 | 0.051 |
| | | yes | 0.02 | 0.049 | 0.047 |
| | Diabetes | no | 0.07 | 0.091 | 0.094 |
| | | yes | 0.075 | 0.093 | 0.094 |
| | FICO | no | 0.052 | 0.093 | 0.103 |
| | | yes | 0.046 | 0.081 | 0.086 |
| | German credit | no | 0.048 | 0.074 | 0.079 |
| | | yes | 0.047 | 0.074 | 0.076 |
| SVM | Adult | no | 0.029 | 0.058 | 0.071 |
| | | yes | 0.035 | 0.06 | 0.068 |
| | Compas | no | 0.015 | 0.033 | 0.031 |
| | | yes | 0.014 | 0.027 | 0.026 |
| | Diabetes | no | 0.076 | 0.101 | 0.097 |
| | | yes | 0.078 | 0.096 | 0.098 |
| | FICO | no | 0.035 | 0.045 | 0.047 |
| | | yes | 0.036 | 0.044 | 0.047 |
| | German credit | no | 0.067 | 0.087 | 0.087 |
| | | yes | 0.063 | 0.091 | 0.08 |
| Mean | | no | 0.042 | 0.068 | 0.07 |
| | | yes | 0.042 | 0.066 | 0.068 |
| | | overall | 0.042 | 0.067 | 0.069 |

**5.3 Comparing the Solutions Returned by the Lex-EAs and Par-EA from a Holistic Perspective (Considering the Four Objectives as a Whole)**

The previous Section compared the results of the two Lex-EAs against the results of Par-EA by considering their ability to optimise each of the four objectives separately. This current section performs instead a holistic comparison between the solutions returned by these two types of EAs, by comparing the overall quality of their returned solutions considering the four objectives as a whole, consistent with the spirit of a genuine multi-objective optimisation approach. Since we are essentially comparing EAs based on two different multi-objective optimisation approaches, i.e. lexicographic optimisation and Pareto dominance, we perform two types of this holistic, multi-objective evaluation: one where the solutions returned by the two types of EAs are compared in terms of Pareto dominance, and another where the solutions are compared in terms of lexicographic optimisation.

Table 8 reports the results of comparing each solution returned by the Lex-EAs against each solution returned by Par-EA, from the perspective of Pareto dominance. That is, for each pair of solutions where one was returned by Lex-EA-1 or Lex-EA-2 and the other was returned by Par-EA, the system checks if one of the solutions in that pair Pareto-dominates the other. These dominance results are reported in the last two columns of Table 8 in the format (W; L; T), denoting the number of wins, losses and ties (in terms of Pareto dominance) for the first EA in the column heading against the second EA in the column heading. For example, for the first row of results in that table, the results in the column "Lex-EA-1 vs Par-EA" show that the solutions returned by Lex-EA-1 Pareto-dominate the solutions returned by Par-EA in 44 cases, the converse was true in 620 cases, and in the other 1,762 cases there was a tie, i.e. none of the two solutions being compared dominated the other.

As shown in the last row of Table 8, with the mean percentages of wins, losses and ties across the 30 experiments, it is clear that in the vast majority (about 90-91%) of the cases, the solutions returned by each Lex-EA and by Par-EA are nondominated with respect to each other. In addition, a solution returned by a Lex-EA is Pareto-dominated by a solution returned by Par-EA in about 8-9% of the cases, whilst the converse (a Lex-EA solution Pareto-dominated a Par-EA solution) occurred in only 0.5% of the cases. The results are very similar for both Lex-EA versions.

This pattern of results is also broadly the same when analysing separately the mean results for the two EA versions with or without resilience integrated into the validity objective, as reported in the last-but-one and last-but-two rows of Table 8. A small difference between the means for these two scenarios is that the use of resilience has somewhat reduced the percentage of cases where a Lex-EA solution was Pareto-dominated by a Par-EA solution, reducing it from 11.4% to 7.7% for Lex-EA-1 and from 10.1% to 6.7% for Lex-EA-2. This reduction was mainly associated with a small increase in the number of ties, rather than an increase in the number of wins for the Lex-EAs.

This reduction can be explained by the fact that integrating resilience into the validity objective has helped the Lex-EAs to return solutions with better values of this objective, resulting in a higher proportion of valid counterfactuals returned by the Lex-EAs (see Table 3); and this made it more difficult for a solution returned by Par-EA to dominate a solution returned by a Lex-EA.

In summary, in the vast majority of the cases the solutions returned by the Lex-EAs and the solutions returned by Par-EA are nondominated with respect to each other; and when a dominance is observed, it is more often the case that a solution returned by Par-EA Pareto-dominates a solution returned by the Lex-EAs. This pattern of results is not very surprising, considering that Par-EA optimises objectives using specifically the Pareto dominance criterion; unlike the Lex-EAs. What could be considered somewhat surprising is that, despite this natural advantage of Par-EA over the Lex-EAs when their returned solutions are compared based on Pareto dominance, in the vast majority of cases there is a tie, i.e. nondominance. However, this can be explained by the fact that, when using Pareto dominance, the percentage of nondominated solutions tends to increase fast with the increase in the number of objectives, and four objectives is already enough to lead to many nondominance results, in these experiments.

Whilst Table 8 compared the Lex-EAs against Par-EA from the perspective of Pareto dominance, to be fair, it is also important to compare Lex-EAs against Par-EA from the perspective of lexicographic optimisation. This is performed in Table 9, where for each pair of solutions where one was returned by Lex-EA-1 or Lex-EA-2 and the other was returned by Par-EA, the system determines which of those two solutions is lexicographic better than the other – i.e., which solution wins the lexicographic tournament (as specified in Algorithm 1); or determines that there is a tie. Note that in this comparison ties are very rare, because they occur only when the two solutions being compared have exactly the same values for all four objectives being optimised. The results in Table 9 are also reported in the format (W; L; T), denoting the number of lexicographic wins, losses and ties for the first EA in the column heading against the second EA in the column heading.

As shown in the last row of Table 9, with the mean percentages of wins, losses and ties across the 30 experiments, it is clear that the Lex-EAs lexicographically win over Par-EA in the large majority of the cases, namely in about 82% of the cases for Lex-EA-1 and about 88% of the cases for Lex-EA-2. As expected, there is a very small proportion of ties (only 0.2%). This same type of pattern is observed when analysing separately the mean percentages for the two scenarios of EAs with or without resilience, as reported in the last-but-one and last-but-two rows of Table 9. Broadly speaking, integrating reliance into the validity objective has somewhat increased the percentages of wins for the Lex-EAs, particularly for Lex-EA-1, whose percentage of wins with resilience was 85.3%, vs. 77.8% without resilience. The increase in percentage of wins with resilience was small for Lex-EA-2.

In summary, in the large majority of the cases the solutions returned by the Lex-EAs are better, in terms of lexicographic optimisation, than the solutions returned by Par-EA. This pattern of results is again consistent with what we should expect, considering that the Lex-EAs optimise objectives using specifically the lexicographic optimisation criterion; unlike Par-EA.

**Table 8:** Comparing Par-EA and Lex-EAs based on Pareto dominance. Each cell in the last two columns shows the results in Win-Loss-Tie format (W; L; T) for each of the Lex-EA's solutions (for each experimental setting), where the result is computed by using *the Pareto dominance criterion* to compare that Lex-EA-returned solution against all the solutions produced by Par-EA (for the corresponding experimental setting)

| Type of model | Dataset | EA with resilience? | Lex-EA-1 vs Par-EA (W; L; T) | Lex-EA-2 vs Par-EA (W; L; T) |
|---|---|---|---|---|
| Neural network | Adult | no | 44; 620; 1,762 | 17; 615; 1,794 |
| | | yes | 3; 94; 3,151 | 5; 245; 2,998 |
| | Compas | no | 6; 39; 1,125 | 17; 30; 1,134 |
| | | yes | 1; 47; 1,277 | 5; 48; 1,277 |
| | Diabetes | no | 0; 934; 3,755 | 8; 600; 4,081 |
| | | yes | 4; 592; 4,484 | 14; 584; 4,482 |
| | FICO | no | 20; 167; 1,104 | 2; 109; 1,180 |
| | | yes | 21; 127; 1,220 | 15; 61; 1,292 |
| | German credit | no | 0; 92; 2,263 | 2; 164; 2,189 |
| | | yes | 4; 224; 2,975 | 2; 290; 2,911 |
| Random Forest | Adult | no | 2; 587; 2,817 | 2; 574; 2,830 |
| | | yes | 3; 478; 2,886 | 9; 182; 3,176 |
| | Compas | no | 2; 34; 828 | 2; 39; 821 |
| | | yes | 5; 33; 1,263 | 9; 32; 1,265 |
| | Diabetes | no | 4; 544; 3,896 | 7; 383; 4,054 |
| | | yes | 6; 251; 5,625 | 9; 358; 5,515 |
| | FICO | no | 18; 219; 1,221 | 34; 197; 1,227 |
| | | yes | 5; 278; 1,599 | 5; 220; 1,657 |
| | German credit | no | 21; 357; 2,962 | 1; 468; 2,871 |
| | | yes | 8; 111; 3,245 | 7; 115; 3,242 |
| SVM | Adult | no | 57; 363; 2,624 | 67; 342; 2,635 |
| | | yes | 159; 678; 3,190 | 51; 346; 3,630 |
| | Compas | no | 10; 81; 2,226 | 67; 37; 2,276 |
| | | yes | 5; 67; 2,942 | 51; 69; 2,941 |
| | Diabetes | no | 3; 689; 4,658 | 18; 634; 4,698 |
| | | yes | 1; 678; 6,037 | 1; 389; 6,326 |
| | FICO | no | 22; 205; 4,375 | 23; 177; 4,402 |
| | | yes | 14; 248; 6,010 | 12; 388; 5,872 |
| | German credit | no | 8; 39; 2,935 | 3; 45; 2,934 |
| | | yes | 66; 319; 4,132 | 0; 314; 4,203 |
| Mean (%) | | no | 0.5%; 11.4%; 88.1% | 0.6%; 10.1%; 89.3% |
| | | yes | 0.6%; 7.7%; 91.8% | 0.3%; 6.7%; 93% |
| | | overall | 0.5%; 9.3%; 90.2% | 0.5%; 8.2%; 91.3% |

Table 9: Comparing Par-EA and Lex-EAs based on lexicographic optimisation. Each cell in the last two columns shows the results in Win-Loss-Tie format (W; L; T) for each of the Lex-EA's solutions (for each experimental setting), where the result is computed by using *the lexicographic-optimisation criterion* to compare that Lex-EA-returned solution against all the solutions produced by Par-EA (for the corresponding experimental setting)

| Type of model | Dataset | EA with resilience? | Lex-EA-1 vs Par-EA (W; L; T) | Lex-EA-2 vs Par-EA (W; L; T) |
|---|---|---|---|---|
| Neural network | Adult | no | 1764; 661; 1 | 1783; 641; 2 |
| | | yes | 2860; 385; 3 | 2997; 249; 2 |
| | Compas | no | 1031; 119; 20 | 1007; 142; 21 |
| | | yes | 1127; 177; 21 | 1105; 196; 24 |
| | Diabetes | no | 3074; 1614; 1 | 3952; 737; 0 |
| | | yes | 4022; 1058; 0 | 4247; 833; 0 |
| | FICO | no | 1062; 223; 6 | 1158; 126; 7 |
| | | yes | 1206; 155; 7 | 1287; 74; 7 |
| | German credit | no | 2010; 333; 12 | 2120; 222; 13 |
| | | yes | 2789; 401; 13 | 2888; 305; 10 |
| Random Forest | Adult | no | 2485; 919; 2 | 2963; 440; 3 |
| | | yes | 2592; 775; 0 | 3187; 180; 0 |
| | Compas | no | 785; 64; 15 | 796; 58; 10 |
| | | yes | 1104; 183; 14 | 1220; 65; 16 |
| | Diabetes | no | 3116; 1326; 2 | 3852; 591; 1 |
| | | yes | 5032; 849; 1 | 5250; 630; 2 |
| | FICO | no | 1100; 353; 5 | 1186; 270; 2 |
| | | yes | 1477; 405; 0 | 1648; 234; 0 |
| | German credit | no | 2716; 618; 6 | 2887; 448; 5 |
| | | yes | 3187; 169; 8 | 3183; 174; 7 |
| SVM | Adult | no | 2199; 845; 0 | 2644; 399; 1 |
| | | yes | 3116; 910; 1 | 3499; 527; 1 |
| | Compas | no | 2029; 270; 18 | 2059; 234; 24 |
| | | yes | 2726; 266; 22 | 2737; 252; 25 |
| | Diabetes | no | 4173; 1174; 3 | 4587; 761; 2 |
| | | yes | 5706; 1006; 4 | 6140; 575; 1 |
| | FICO | no | 3702; 895; 5 | 3808; 789; 5 |
| | | yes | 5410; 857; 5 | 4997; 1271; 4 |
| | German credit | no | 2804; 159; 19 | 2872; 92; 18 |
| | | yes | 3931; 582; 4 | 4025; 485; 7 |
| Mean | | no | 77.8%; 21.9%; 0.3% | 86.1%; 13.6%; 0.3% |
| | | yes | 85.3%; 14.5%; 0.2% | 88.7%; 11.1%; 0.2% |
| | | overall | 81.7%; 18.1%; 0.2% | 87.6%; 12.2%; 0.2% |

## 6 Conclusions

This work proposed a novel Evolutionary Algorithm (EA) for generating counterfactuals as a form of explanation for the predictions of a black-box classification model. Counterfactuals are a particularly promising form of explanation because they are designed to be actionable, by explicitly indicating how some feature value(s) in an instance should be changed in order for a black-box model to switch its prediction for that instance from an undesirable negative class (e.g. denying a loan) to a desirable positive class (e.g. granting a loan). The proposed EA performs multi-objective optimisation, optimising four objectives (counterfactual properties), namely: (a) a counterfactual's validity, i.e., its ability to predict the desirable class; (b) the distance between a counterfactual and the point of interest (instance whose classification is being explained by the counterfactual); (c) a counterfactual's sparsity, i.e., the number of features whose values changed in the counterfactual; and (d) a counterfactual's plausibility, i.e. the distance between the counterfactual and its nearest training instance.

This work's two main contributions are as follows. First, it proposed a novel type of multi-objective EA for counterfactual generation based on lexicographic optimisation, rather than based on Pareto dominance as in the most related work on EA for counterfactual generation in the literature [Dandl et al. 2020]. The lexicographic approach consists of optimising objectives in a predefined order of priority. In this work, the use of this approach allowed us to incorporate into the EA the background knowledge that, among the four counterfactual properties (objectives) being optimised by the EA, validity is the highest-priority objective, since strictly speaking an invalid counterfactual is not a counterfactual at all.

As the second contribution, this work proposed an extension to the definition of the objective of validity, based on measuring the resilience of a counterfactual to violations of monotonicity constraints which are intuitively expected by users (as discussed in Section 3.2). Satisfying monotonicity constraints is often an effective approach for improving the interpretability of classification models [Cano et al. 2019], but this approach has been very under-explored in the literature on counterfactual generation in general, and to the best of our knowledge there is no previous EA for counterfactual generation that uses monotonicity constraints to improve a counterfactual's quality. In this work, the monotonicity constraint-based extension of the validity objective has been incorporated into both the proposed lexicographic optimisation-based EA (Lex-EA) and the Pareto dominance-based EA (Par-EA) described in [Dandl et al. 2020].

Lex-EA has been compared with Par-EA in 15 experimental settings – consisting of learning 3 types of black-box classification models (neural networks, random forests and Support Vector Machines (SVM)) from each of 5 different classification datasets. Overall, the computational results have shown that the proposed Lex-EA has clearly outperformed Par-EA regarding their ability to optimise the validity objective, with Lex-EA achieving a considerably higher percentage of valid counterfactuals than Par-EA, among all counterfactuals returned by each type of EA. This was an expected result, since Lex-EA was explicitly designed to give the highest priority to the optimisation of this objective. In addition, an important and somewhat unexpected result is that Lex-EA still managed to be competitive with Par-EA regarding their ability to optimise the three other aforementioned objectives, with both types of EAs achieving on average broadly similar values for those three objectives in their returned counterfactuals.

In addition, the proposed incorporation of a measure of the resilience of a counterfactual to violations of monotonicity constraints into the validity objective has led to a substantial increase in the validity of the counterfactuals generated by Lex-EA, and also to a small increase in the validity of the counterfactuals generated by Par-EA.

Two suggested avenues for future research are as follows. First, in this work the same fixed value of the tolerance threshold parameter (in the lexicographic tournament selection procedure) was used for all objectives being optimised. This fixed parameter value has led Lex-EA to achieve good results as reported earlier, but in principle the results for Lex-EA could potentially be improved, in future work, by optimising this lexicographic optimisation-related parameter value for each objective (e.g. using grid search); although this would lead to considerably longer EA running times.

Second, the proposed approach for measuring a counterfactual's resilience to violations of monotonicity constraints has focused on the most commonly used type of monotonicity constraints, involving a pair of numeric variables (a numeric feature and the probability of the positive class, in this work). In future work, it would be interesting to extend this approach for coping with categorical features and monotonicity constraints involving multiple features and the positive-class probability.

**Acknowledgement**
We would like to thank Susanne Dandl for providing assistance in understanding and using the program code of the MOC Evolutionary Algorithm.

**Appendix**

Table A.1 lists the non-actionable features for each dataset. As described in the main text, these are features that cannot be mutated for creating a counterfactual.

**Table A.1:** Non-actional features for each dataset

| Dataset | List of non-actionable features |
|---|---|
| Adult Census Income | age, education, marital_status, relationship, race, sex, native_country |
| Compas | age, age_cat, race, sex |
| Diabetes | age, pregnancies |
| FICO | externalRiskEstimate |
| German Credit | age, sex |